\pdfoutput=1

\documentclass[11pt]{article}

\usepackage[preprint]{acl}

\usepackage{times}
\usepackage{latexsym}
\usepackage{pifont}
\usepackage{amssymb}

\usepackage[T1]{fontenc}

\usepackage[utf8]{inputenc}

\usepackage{microtype}

\usepackage{inconsolata}

\usepackage{amsmath,amsfonts,bm}

\def\eqref#1{equation~\ref{#1}}

\def\1{\bm{1}}

\DeclareMathAlphabet{\mathsfit}{\encodingdefault}{\sfdefault}{m}{sl}
\SetMathAlphabet{\mathsfit}{bold}{\encodingdefault}{\sfdefault}{bx}{n}

\usepackage{hyperref}
\usepackage{url}
\usepackage{graphicx}
\usepackage{amsmath}
\usepackage{multirow}
\usepackage{capt-of}
\usepackage{wrapfig}

\newcommand{\AlgName}{LexiContrastive Grounding~}
\newcommand{\AlgNameNE}{LexiContrastive Grounding}
\newcommand{\AlgNameAb}{LCG~}
\newcommand{\AlgNameAbNE}{LCG}
\newcommand{\AlgVokenName}{LexiVoken Grounding~}
\newcommand{\AlgVokenNameNE}{LexiVoken Grounding}

\title{Lexicon-Level Contrastive Visual-Grounding Improves Language Modeling}

\author{Chengxu Zhuang$^1$ \and Evelina Fedorenko$^{1, 2, 3}$ \and Jacob Andreas$^4$ \\
$^1$McGovern Institute for Brain Research, MIT\\
$^2$Department of Brain and Cognitive Sciences, MIT\\
$^3$The Program in Speech and Hearing Bioscience and Technology, Harvard University\\
$^4$CSAIL, MIT\\
\texttt{\{chengxuz,evelina9,jda\}@mit.edu} \\
}

\begin{document}
\maketitle

\begin{abstract}
Today's most accurate language models are trained on orders of magnitude more language data than human language learners receive---but with no supervision from other sensory modalities that play a crucial role in human learning.
Can we make LMs' representations and predictions more accurate (and more human-like) with more ecologically plausible supervision? 
This paper describes \AlgNameNE{} (LCG), a grounded language learning procedure that leverages visual supervision to improve textual representations.
\AlgName combines a next-token prediction strategy with a contrastive visual grounding objective, focusing on early-layer representations that encode lexical information.
Across multiple word-learning and sentence-understanding benchmarks, \AlgName not only outperforms standard language-only models in learning efficiency, but also improves upon vision-and-language learning procedures including CLIP, GIT, Flamingo, and Vokenization.
Moreover, \AlgName improves perplexity by around 5\% on multiple language modeling tasks.
This work underscores the potential of incorporating visual grounding into language models, aligning more closely with the multimodal nature of human language acquisition.
\end{abstract}

\section{Introductions}

Neural language models (LMs; \citealp{BERT, RoBERTa, GPT2, GPT3}) have shown utility in modeling aspects of human language processing, including neuronal responses to linguistic stimuli \citep{schrimpf2021neural, caucheteux2022brains, goldstein2022shared} and data about human language production \citep{arehalli2020neural} and comprehension \citep{wilcox2020predictive}. Nevertheless, these models currently lack plausibility as models of human cognitive development. This discrepancy primarily stems from the immense volume of training data necessitated for effective LM performance, surpassing---by orders of magnitude---the linguistic input received during human language acquisition~\citep{zhang2020you, warstadt2022artificial}. Specifically, children may be exposed to at most sixty million words in their first five years \cite{frank2023bridging}, whereas training modern LMs requires hundreds of billions of words. Can insights from human language acquisition guide the training of new LMs that are both better cognitive models and more sample-efficient in an absolute sense?

\begin{figure*}[!t]
\centering
\includegraphics[width=\textwidth] {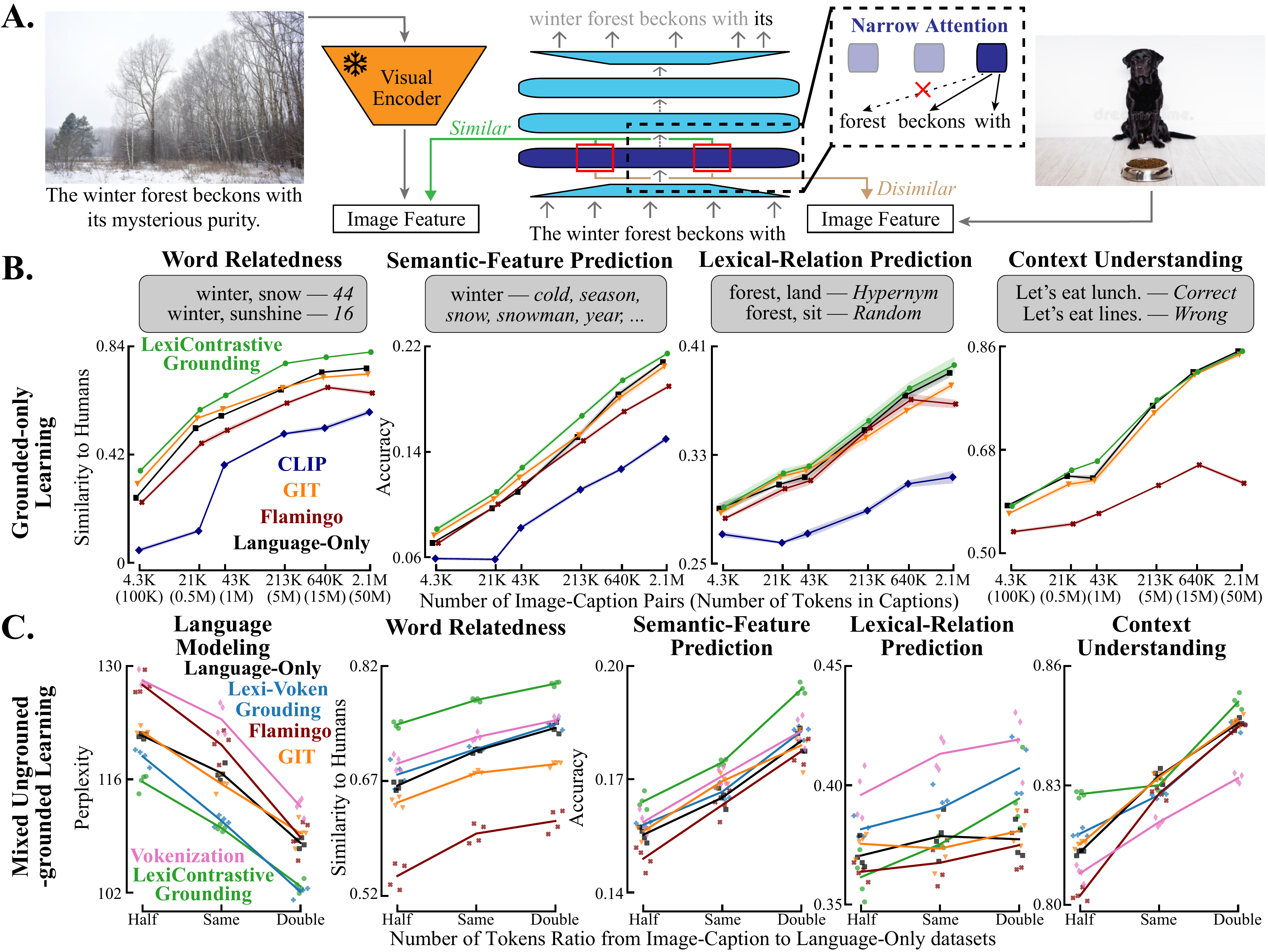}
\captionof{figure}{
\textbf{\AlgName models leverage visual information to facilitate word learning when they are trained on image-caption datasets.}
\textbf{A.} Pretraining schema for the \AlgName models. The images are sent to a frozen visual encoder pretrained using unsupervised learning algorithms to generate image features. These image features and the hidden representations of the first layer after the token-embedding layer are used to compute a vision-language contrastive loss. This loss is added to the next-token prediction loss to form the final loss.
\textbf{B.} Results from the grounded-only learning scenario on word-learning benchmarks for \AlgName (\ding{108}), Language-Only (\ding{110}), CLIP (\ding{117}), GIT (\ding{116}), and Flamingo (\ding{54}). 
The X-axis is plotted in the log scale. Each point represents the average performance from four models initialized from different random seeds, and the line width represents the S.E.M. from these four models.
\textbf{C.} Results from the mixed-learning scenario on the language modeling and the word learning benchmarks. We also add \AlgVokenName (\ding{58}) and Vokenization (\ding{169}) models. 
The ungrounded dataset is Smashwords-5M. 
Different dots of the same color represent models with different random initialization seeds.
}
\label{fig_main}
\end{figure*}

A key contrast in language learning between humans and LMs is that humans ground language learning in perceptual signals across various modalities, encompassing hearing, touch, and vision \citep{clerkin2017real, west2017language, seidl2023touch, schroer2023looking}.
Multi-modal training has also been studied in natural language processing as a potential avenue towards more human-like language learning~\citep{bisk2020experience}.
Encouragingly, recent years have witnessed a surge in the development of multi-modal models and learning algorithms, primarily tailored for tasks requiring simultaneous reasoning across both modalities~\citep{CLIP, GIT, Flamingo, FLAVA, unified_io, Vokenization}.
However, none of these multi-modal models have been shown to learn language more efficiently than language-only models. 
In fact, \citet{zhuang2023visual} show that several different visual-language models (CLIP, GIT, and Flamingo) learn word meanings \emph{less} efficiently than models trained on language alone. This finding suggests that these existing visual-language learning algorithms cannot model how humans leverage vision to help learn language.
However, \citet{zhuang2023visual} also find that vision-language contrastive learning \citep{CLIP} with images and \emph{single words} yields representations that are sometimes comparably well aligned with human judgments, but qualitatively different from, representations learned from text alone.

Inspired by these findings, we propose a new visually grounded language learning procedure we call \AlgName (\AlgNameAbNE),
which combines the next-token prediction objective and a word-level contrastive visual grounding objective.
Crucially, we apply this contrastive objective to the early-layer representations in the LM, as these representations are closer to containing only lexicon-level information. 

Unlike the methods studied by \citet{zhuang2023visual}, we find that \AlgName yields improved learning efficiency compared to language-only models.
We evaluate it, and other visual-language learning procedures, on two learning scenarios: grounded-only and mixed ungrounded-grounded language learning. In both scenarios, we systematically vary the size of the training datasets. Additionally, we vary the source of the ungrounded corpus in the mixed-learning scenario. After training the models, we evaluate them on tasks that assess different aspects of word and language learning, including semantic similarity judgment, lexical relation and semantic norm prediction, and language modeling. \AlgName outperforms existing methods on most of the evaluated benchmarks, demonstrating a consistent and significant benefit of visual grounding on language modeling compared to the language-only models.

To the best of our knowledge, \AlgName is the first multi-modality learning algorithm to transfer benefits from visual grounding to (unconditional) language modeling. 
These results show how natural language semantics can be better acquired by grounded learning, and suggest steps toward human-level efficiency in language learning with LMs.

\section{Background}\label{sec_rev}


\textbf{Grounded language learning algorithms in AI.}
In recent years, there have been notable advancements in multi-modality learning. For instance, the Vokenization model \citep{Vokenization} finds contextually relevant images (called “vokens”) for language tokens and uses the vokens to additionally supervise the LMs. Although the Vokenization model is shown to outperform language-only baselines on language-understanding benchmarks like MNLI \citep{MNLI}, the model is first finetuned on the corresponding training sets of these benchmarks. This finetuning process makes it unclear whether the learned representation is directly better than the representations of the language-only models. The CLIP model produces transferable visual representations and word representations that demonstrate strong performance on certain tests of word similarity~\citep{CLIP, WordSimBench}, after being trained contrastively on a massive amount of image-caption pairs. In contrast, GIT is a generative model that achieves state-of-the-art performance on various visual-language tasks, such as image captioning and visual question answering, through utilizing visual inputs to condition next-word predictions \citep{GIT}. The related Flamingo model also yields strong results on these tasks by employing visual representations to modulate attention in a transformer language model \citep{Flamingo}. 

\textbf{Models of human language acquisition using ungrounded and grounded learning algorithms.}
\citet{BabyBERTa} and \citet{BabyLMFindings} target grammar learning in models trained on small ungrounded datasets. \citet{chang2022word} analyze word-acquisition trajectories in language-only models, but their focus on model surprisal changes during training makes their findings less relevant to learning word meanings. For grounded models, \citet{berger2022computational} and \citet{portelance2023learning} propose multi-modality algorithms for understanding word acquisition, with specific emphasis on word categories and function words, respectively.
\citet{Lake_science_paper} apply a CLIP-like learning algorithm on first-person videos collected from children and report that the trained visual-language model yields good performance in visual-referent mapping tasks on the same child’s experience and modest generalization ability to out-of-domain datasets.
\citet{zhuang2023visual} evaluate multiple existing visual-language models on word-learning benchmarks and show that visual-grounding yields limited and conditioned help in low-data regimes.
Our work introduces a stronger visual-language learning algorithm that outperforms these existing algorithms in different learning scenarios.

\section{\AlgNameNE}\label{sec_alg}

\citet{zhuang2023visual} find that vision-and-language contrastive learning \citep{CLIP} applied to images and individual words within corresponding captions produces surprisingly high-quality word representations. 
Indeed, across multiple word-learning benchmarks, \citet{zhuang2023visual} report that this image-word contrastive learning objective is more efficient than the next-token prediction objective on learning to relate words in a human-like way and to predict their semantic features.
Moreover, they find that having more context in the linguistic input than a single word yields lower efficiency in learning word meanings, with the image-sentence contrastive learning objective being significantly less efficient compared to both the image-word and the language-only objective.
More importantly, the image-word and language-only representations differ from each other as only the one learned with visual grounding encodes the meanings of concrete words in a more human-like way compared to less concrete words.
This distinction indicates that a stronger model might be produced by combining these two learning algorithms in some way.

\textbf{Intuition:}
Motivated by this, we first compute a cross-modality contrastive learning loss on representations from the first hidden layer of the model (see Fig.~\ref{fig_main}\textbf{A}).
We select the representations of the first layer because they merge less information from the linguistic context.
To enforce these representations to encode even less context, we further limit the attention operation of the first layer to only attend to the previous two tokens.
The cross-modality contrastive loss is then computed from all the token-level representations from the first layer of all examples within a batch. 
This loss differs from the loss computed by CLIP because CLIP extracts a single sentence-level representation for the whole caption while we use all available token-level representations from the early layer.
We then linearly mix this contrastive loss with the next-token prediction loss on the whole caption to get the final loss (see Fig.~\ref{fig_main}\textbf{A}).

\textbf{Objective function:}
To be more concrete, let $(v_i, c_i)$ denote the pairs of image ($v_i$) and caption ($c_i$) within one batch, where $i$ ranges from 1 to batch size $n$ and $c_i$ contains $m_i$ tokens: $(t_{i, 1}, t_{i, 2}, …, t_{i, m_i})$.
We then use $f_{L}^1(c_i)$ to represent the output of the first hidden layer of the neural language model.
Because $f_{L}^1(c_i)$ is a matrix of shape $m_i \times d$, where $d$ is the dimension for the hidden representation in the model, we define $f_{L}^1(c_i, j)$ to be the $j$-th column of this matrix.
Assuming that $f_{V}(v_i)$ represents the visual feature computed from a visual encoder $f_{V}$, the matching score $s(i, j, k)$ between $v_i$ and $f_{L}^1(c_k, j)$ is then defined as:
\begin{equation*}
    s(i, j, k) = (M_{V}f_{V}(v_i))^{\top} M_{L}f_{L}^1(c_k, j)  / \tau
\end{equation*}
$\tau$ is a trainable positive value, while $M_V$ and $M_L$ are $d\times d$ matrices.
The contrastive loss is then:
\begin{equation*}
\begin{aligned}
&\mathcal{L}_{c} = \sum_{i=1}^{n} \sum_{j=1}^{m_i} \frac{1}{2} \frac{e^{s(i, j, i)}}{\sum_{k=1}^{n} e^{s(k, j, i)}} + \frac{1}{2} \frac{e^{s(i, j, i)}} {\mathrm{neg}(i, j)} \\ 
&\mathrm{neg}(i, j) = e^{s(i, j, i)} + \sum_{k=1}^n \sum_{o=1}^{m_k} (1 - \delta_i(k)) e^{s(i, o, k)}
\end{aligned}
\end{equation*}
where $\delta_i(k)$ is 1 when $k=i$ and 0 otherwise.
The final loss on this grounded batch is then:
\begin{equation*}
\mathcal{L}_{g} = \lambda_c \mathcal{L}_{c} + \mathcal{L}_{l}
\end{equation*}
where $\mathcal{L}_{l}$ represents the next-token prediction loss on the captions.

\textbf{Visual encoder.}
Following \citet{zhuang2023visual}, we use a Vision Transformer (ViT; \citealp{ViT}) pretrained on unlabeled ImageNet images using the DINO algorithm \citep{DINO}, which is a strong unsupervised visual learning algorithm. The image feature is the hidden representation at the \verb|[CLS]| token from the last layer.

\section{Experiment setup}

\subsection{Training Datasets}
The grounded datasets contain image-caption pairs from the Conceptual-Captions-12M dataset \citep{CC12M}.
We only used images that were valid in August of 2022.
In the mixed learning scenario, we use samples from the Smashwords containing 5M and 15M tokens as well as a subset of CHILDES \citep{CHILDES} containing 5M tokens as the ungrounded dataset.
These three ungrounded datasets cover widely available corpus comprising a significant part of the training materials of high-performing LMs (Smashwords) and more development-relevant corpus (CHILDES).
The training in the mixed learning scenario simultaneously draws two batches from both the ungrounded and grounded datasets and optimizes a linear mix of the two losses.
The mixing weight is varied across multiple choices and decided based on the perplexity measure on the validation set of the ungrounded dataset.

\subsection{Evaluation benchmarks}

We evaluate our models on four of the word-learning evaluation benchmarks proposed by \citet{zhuang2023visual}: Word Relatedness, Semantic Feature Prediction, Lexical Relation Prediction, and Context Understanding benchmarks. Our selection covers lexical-level and sentence-level evaluations.
This selection includes benchmarks where visual grounding has shown benefits in low-data situations, specifically the Word Relatedness and Semantic Feature Prediction benchmarks, as well as two other benchmarks where grounding has not proven to be helpful.
Additionally, we also evaluate the perplexity of the trained models in the mixed-learning scenario on the held-out test set of the ungrounded dataset.
Since these benchmarks are proposed by \citet{zhuang2023visual}, we only briefly introduce them here.

\textbf{\textit{Lexical-level:} Word Relatedness Benchmark.}
This benchmark evaluates model performance on predicting human word relatedness judgments using MEN \cite{MTest3000}, focusing on semantic similarities between word pairs (see Fig.~\ref{fig_main}\textbf{B}). Models are assessed by extracting word representations, calculating pairwise cosine similarities, and then comparing these to human judgments via Spearman correlations to identify the optimal layer.

\textbf{\textit{Lexical-level:} Semantic Feature Prediction Benchmark.}
This benchmark evaluates LMs through semantic norm prediction tasks, using a dataset by \citet{Buchanan2019}, where human annotators list features of words (see examples in Fig.~\ref{fig_main}\textbf{B}). The evaluation involves training a linear probe on model-derived word representations to predict these features, selecting the best layer based on validation set accuracy, and reporting its performance on a separate test set.

\textbf{\textit{Lexical-level:} Lexical Relation Prediction Benchmark.}
This benchmark assesses the ability of models to accurately predict complex word relationships (such as synonyms, hyponyms, etc.) using the CogALex-V dataset \citep{CogALexV}. This dataset comprises over 2500 word pairs in both training and test sets, to evaluate model precision in classifying word pairs into five categories: synonymy, antonymy, hypernymy, part-whole meronymy, and random (see examples in Fig.~\ref{fig_main}\textbf{B}). A similar linear probing method is applied here to the difference between two word representations to predict the lexical relations.

\textbf{\textit{Sentence-level:} Context Understanding Benchmark.}
This benchmark tests if models can discern appropriate contexts for word usage. Its creation involves selecting real sentences featuring target words from online sources, then altering these sentences to create inappropriate contexts for the words (see examples in Fig.~\ref{fig_main}\textbf{B}). Model performance is evaluated based on their ability to correctly identify the original sentence as more probable than its modified counterpart. The original benchmark created by \citet{zhuang2023visual} generates sentence pairs for nouns, verbs, and adjectives, which divides the benchmark into three sub-benchmarks. We report the average performance across these three sub-benchmarks.

\textbf{\textit{Language Modeling:} Perplexity.}
We evaluate the perplexity measure on the held-out sets of the ungrounded datasets in the mixed-learning scenario.
The perplexity is measured for words in the sequence with at least 64 and at most 127 prior tokens as the context.
Because all our models use the same tokenizer, their perplexity measures are directly comparable.

\subsection{Baselines}
\textbf{Language-Only Models.}
We train these models only using image captions or the ungrounded input. The training uses the next-token prediction objective function. We use a six-layer variant of the GPT-2 architecture \citep{GPT2}, following \citet{zhuang2023visual}. 
This shallower architecture performs similarly to its deeper counterpart in the tasks we evaluate \citep{zhuang2023visual}.

\textbf{CLIP.}
We train the CLIP models following the visual-language contrastive learning objective proposed by \citet{CLIP}.
This objective optimizes the language model to produce a caption embedding that is similar to its corresponding image embedding and dissimilar to embeddings of other images.
During training, we freeze the visual encoder to be the DINO-pretrained ViT and train the language model from scratch.
Although we still call the trained models CLIP models, they are not the models pretrained by \citet{CLIP}.
 
\textbf{Flamingo.}
This model utilizes visual representations to modulate attention within the transformer \citep{Flamingo}. This modulation is performed by additional cross-attention layers added between the original self-attention layers. The whole model, including both the cross-attention and the self-attention layers, is then trained from scratch using the next-token prediction objective. When Flamingo is trained on ungrounded input, the cross-attention layers are not used. Although these cross-attention layers contain additional trainable parameters, they are not used during our evaluation since we use language-only input.

\textbf{GIT.}
This algorithm treats the image feature as part of the linguistic context by concatenating it with the output of the word-embedding layer \citep{GIT}. The concatenated representation is then sent to the transformer to perform the same next-token prediction task.

\textbf{Vokenization.}
This algorithm \citep{Vokenization} first trains a contextual token-image matching model on image-caption pairs.
Since the algorithm aims to use visual information to improve text-only language modeling performance, we only test this algorithm in the mixed ungrounded-grounded learning scenario.
We then run this matching model on both ungrounded and grounded datasets to map each of the contextualized token representations to the image that is the most semantically relevant to the representation.
Following \citet{Vokenization}, we choose the image from an independent dataset (Visual Genome; \citealp{VisualGenome}) that is not used in either of the learning scenarios we evaluate.
The index of the selected image is then used as the “voken”.
The final training loss on both ungrounded and grounded datasets is to simultaneously predict the next token as well as the next voken using two readout heads.

\subsection{Training Details}
We set $\lambda_c$ in our algorithm to be 0.3. This choice is supported by ablation studies in Sec. \ref{sec_ablation}.
For the grounded learning scenario, we vary the dataset size from 4.3K to 2.1M image-caption pairs. These pairs are used to train the models for multiple epochs.
Following \citet{zhuang2023visual}, the number of epochs is determined independently for each dataset scale by the loss on the validation set.
For the mixed-learning scenario, we simultaneously train the models on both ungrounded and grounded input.
For each of the three ungrounded datasets (Smashwords-5M, Smashwords-15M, and CHILDES-5M), we vary the size of the corresponding grounded dataset so that it contains either half, the same amount, or double the number of tokens compared to the ungrounded dataset.
This yields nine training setups in total.
In each of these training setups, the loss is a linear mix of two losses computed separately from each input. We vary this mix weight among several candidate values and select the best one according to the perplexity measure on the validation set of the ungrounded dataset. This selection is independently done for each setup and each learning algorithm.

More details can be found in Appendix~\ref{ap_train_eval}.

\section{Results}\label{sec_res}

\subsection{In the grounded-only learning scenario, \AlgNameAb learns word meanings more efficiently than Language-Only models}

We first show the results from the grounded-only learning scenario.
On the Word Relatedness and Semantic Feature Prediction benchmarks, \AlgName models achieve significantly better results on all the dataset scales compared to all the other models, including both Language-Only models and other multi-modality learning models (see Fig. \ref{fig_main}\textbf{B}, left two panels).
Although \citet{zhuang2023visual} also reported benefits from visual grounding on these two benchmarks, their benefits are only in low-data regimes and become smaller in larger datasets, while our \AlgNameAb algorithm yields consistent benefits up to the largest scale.
Even on the other two benchmarks where visual grounding was not found useful by \citet{zhuang2023visual}, \AlgNameAb performs slightly but significantly better than all the other models in small dataset scales (see Fig. \ref{fig_main}\textbf{B}, right two panels).
The improvement on the Context Understanding benchmark also shows that our algorithm better learns not just word-level but also sentence-level representations than other models.
Taken together, these results show that \AlgName effectively leverages visual information to facilitate the learning of word meanings and outperforms both language-only and other visual-language learning algorithms on word learning.
  
\subsection{In the mixed learning scenario, \AlgNameAb improves language modeling across different data sources and scales}

Both humans and models receive ungrounded language input during learning and need to learn in a mixed scenario. 
To explore whether visual grounding helps language learning in this mixed scenario, we train models on both grounded and ungrounded datasets.
This also addresses potential concerns that the GIT and Flamingo models only receive image-caption input in the grounded-only scenario and, therefore, suffer from a domain change when they are tested in language-only evaluation benchmarks.
We additionally test the Vokenization learning algorithm in this mixed scenario because this algorithm was proposed to leverage visual-language alignment to advance the learning on language-only input \citep{Vokenization}.

We find that the \AlgNameAb algorithm achieves better performance than existing algorithms on general language modeling, measured by perplexity in the held-out set of ungrounded datasets (see Fig.~\ref{fig_main}\textbf{C}, the leftmost panel).
This improvement is robust with respect to the sources of the ungrounded dataset and the sizes of both the grounded and ungrounded datasets (see Appendix Fig.~\ref{ap_fig_cotrain}).

Furthermore, we develop an additional algorithm by using the Vokens acquired in the Vokenization process to also ground the lexicon-level representations.
We use this new algorithm, called \AlgVokenNameNE, to verify that the Vokens computed by us encode meaningful signals and support the choice of using cross-modality contrastive learning objective as the grounding objective.
Indeed, we find that the \AlgVokenName models yield better results than Language-Only models on the perplexity measure but still underperform the \AlgName models.

In addition to the improvement in perplexity, \AlgName also illustrates advances in the Word Relatedness and Semantic Feature Prediction benchmarks (see Fig.~\ref{fig_main}\textbf{C}).
Interestingly, the Vokenization and \AlgVokenName models outperform the other models on the Lexical Relation Prediction benchmark, suggesting that the voken signals help encode semantic relations between words.

\begin{figure*}[t]
\centering
\includegraphics[width=0.97\textwidth] {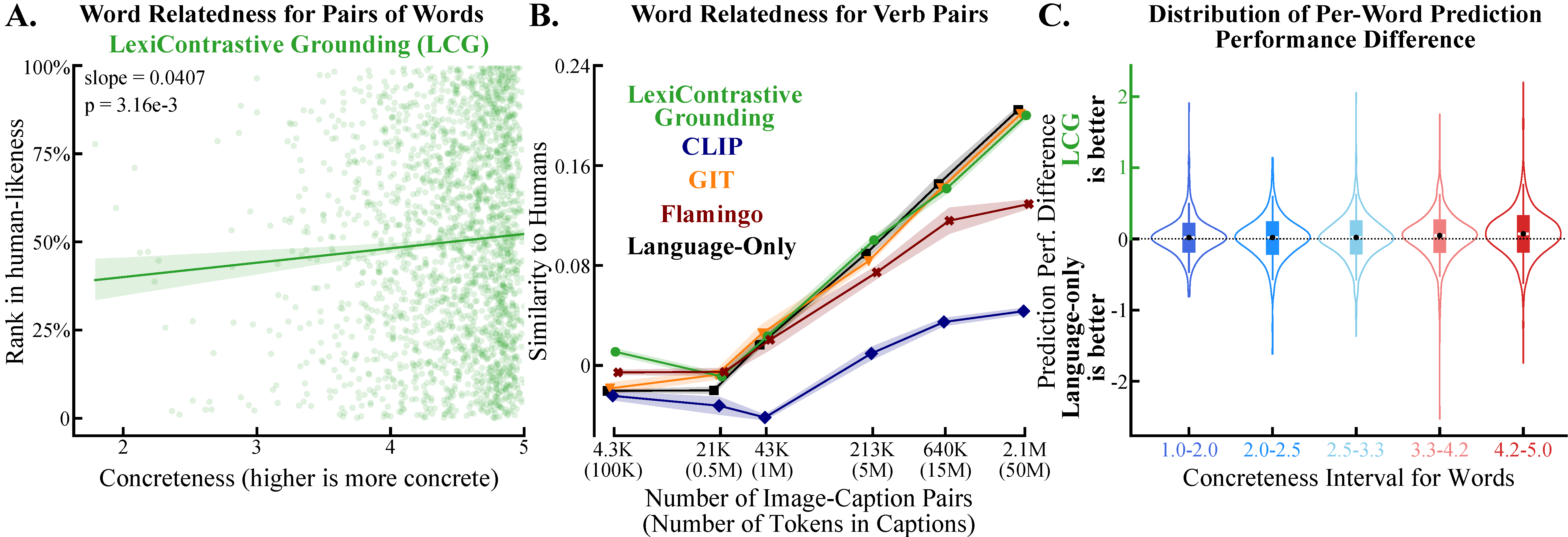}
\captionof{figure}{
\textbf{\AlgName models better learn concrete words than Language-Only models.}
\textbf{A.} Scatter plot for an analysis on the word-relatedness benchmark for the \AlgNameAb model trained with 2.1M image-caption pairs. Each point on the plot corresponds to a pair of words, with its Y-value indicating the relative rank obtained by sorting the word pairs based on the difference between human and model judgments. A greater Y-value signifies a closer resemblance to human judgment. Additionally, linear regression lines are depicted on the graph along with their respective $95\%$ confidence intervals.
\textbf{B.} The results of SimVerb-3500, a word-relatedness benchmark evaluating models only on verb words. The marker-to-model map is the same as that in Fig.~\ref{fig_main}.
\textbf{C.} Distributions of the per-word prediction performance difference between \AlgName and Language-Only models grouped by concreteness of words. The prediction performance is the negative loglikelihood of the corresponding word averaged across all appearances in the test dataset.
The \AlgName and Language-Only models are taken from the “same” condition in Fig.~\ref{fig_main}\textbf{C}. A positive difference means that the \AlgName model is better than the Language-Only model.
}
\label{fig_analysis}
\end{figure*}

\subsection{Concrete words are better learned by \AlgNameAb compared to abstract words}

\citet{zhuang2023visual} show that visual grounding facilitates the learning of concrete words more than abstract words.
Using a similar analysis method, we also find that the \AlgName models trained on 2.1M image-caption pairs relate the concrete words in a more human-like way than the abstract words (see Fig.~\ref{fig_analysis}\textbf{A} and Appendix \ref{ap_analysis}).
Since concreteness does not influence how human-like the Language-Only models relate different words, our finding suggests that the \AlgName models yield more human-like representations because they better acquire the meanings of concrete words compared to the Language-Only models.

\citet{zhuang2023visual} also report that visual grounding contributes little to acquiring relatedness structure in verbs, since the grounding uses visual features from static images, which may only contain limited information about actions. 
After evaluating the Word Relatedness benchmark using a human judgment dataset collected for verb pairs (SimVerb-3500; \citealp{SimVerb3500}), we find that our \AlgName models perform highly similarly to the Language-Only models.
This finding suggests that grounding on more than static images is possibly needed to yield more human-like representations.

Finally, we explore how the better language modeling performance in the mixed learning scenario can be partially explained by the fact that \AlgName models better represent concrete words.
We calculate the averaged performance of next-token predictions for each word across the entire test set.
We then compare the performance of \AlgName and Language-Only models for each word and investigate whether this performance difference is dependent on the concreteness of the word.
As shown in Fig. \ref{fig_analysis}\textbf{C}, concrete words are indeed more accurately predicted by \AlgName models compared to Language-Only models.
This suggests that part of the lower perplexity of \AlgName models in the mixed learning scenario is due to the more human-like representations of concrete words in these models.

\begin{figure*}[t]
\centering
\includegraphics[width=\textwidth] {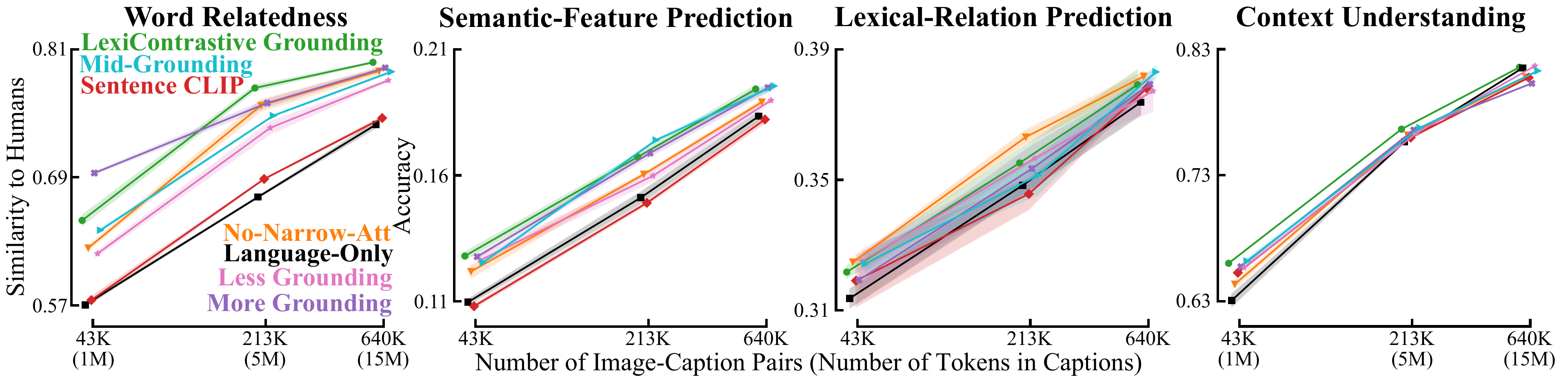}
\captionof{figure}{
\textbf{Ablation studies support the algorithm design of \AlgNameNE.}
Less Grounding (\ding{72}) model changes $\lambda_{c}$ to 0.1. More Grounding (\ding{54}) changes it to 1. No-Narrow-Att (\ding{116}) model has the typical attention layer as the first layer. Mid-Grounding ($\blacktriangleright$) model computes the grounding loss from the third layer. Sentence CLIP (\ding{117}) model computes the sentence-level CLIP loss from the top layer as the grounding loss.
}
\label{fig_ablation}
\end{figure*}

\subsection{Ablation studies support the algorithm design of \AlgNameAbNE}\label{sec_ablation}

To validate the algorithm design of \AlgNameNE, we perform ablation studies on this algorithm and compare the performance of the ablated algorithms to the original performance on the word-learning benchmarks when all of them are trained on grounded-only datasets.
As seen in Fig. \ref{fig_ablation}, \AlgName models achieve generally better results than the other ablated models. 

\section{Discussion}\label{sec_dis}

In this paper, we introduce \AlgNameNE, a visually grounded language learning objective motivated by models of grounded language acquisition in humans.
\AlgNameAb combines next-token prediction with a word-level contrastive visual grounding objective applied to early-layer representations. \AlgNameAb not only outperforms other visual-language learning algorithms across various benchmarks, including evaluations of similarity, lexical relations, and semantic norms, but also surpasses traditional language-only models in learning efficiency on language modeling. This result underscores the potentially significant benefits of visual grounding in language modeling, offering insights into the role of multimodal learning in human-like language acquisition and suggesting a pathway toward more efficient and cognitively aligned language learning technologies.

Our analysis shows that the word meanings acquired by \AlgName are more human-like when the words are concrete.
This concreteness-based bias should not exist for a perfectly human-aligned word-meaning encoding. Therefore, this shows that the representations learned by \AlgName still differ from those in human adults.
One possible explanation for this difference is that an additional learning mechanism is needed to augment the \AlgName algorithm to better learn abstract words.
Another possible explanation is that our training corpus needs to be closer to what human adults perceive in both its quantity and distribution.
Since we only have at most 50M tokens in the training set, our learned representations may better capture what children have compared to what adults have.
More experiments are needed to test these explanations.

Because \AlgName yields benefits on language modeling through leveraging visual grounding, just as (sighted) children leverage visual input during language learning, our approach may be useful as a model of grounded language acquisition in humans.
We note, however, that there may be significant differences between the visual encoder used in our approach and human visual encodings---although the current DINO-pretrained ViT has been shown to be similar to the ventral visual stream of human and non-human primate adults \citep{zhuang2021unsupervised, konkle2022self, zhuang2022well}, it is unclear whether this visual encoder accurately models the visual system of children.
Moreover, the data that this visual encoder is trained on, which is ImageNet \citep{deng2009imagenet} without its labels, is also very different from what children perceive during their development.
Training the visual encoder on datasets like SAYCam \citep{SAYCam} is needed to better capture the development dynamics in children.

\section*{Limitations}\label{sec_lmt}

One limitation of our study is that we only aim to use visual grounding to help learn language but not to predict the coupled text.
Therefore, the \AlgName algorithm likely underperforms the GIT and Flamingo models on image captioning tasks.
To address this, our algorithm needs to be augmented with an extra mechanism to leverage the visual feature to help predict tokens, such as what is implemented in GIT and Flamingo.

Another limitation of our study is that the new algorithm only grounds the lexicon-level representations on visual input, which potentially limits the benefit on syntax learning from visual grounding.
Although it is unclear whether such benefits exist, allowing a potential pathway for the visual features to contribute to syntax learning is an interesting and important next step.

Finally, the visual encoder used in our algorithm is pretrained using DINO and frozen during language learning. \citet{zhuang2023visual} have shown that using a better pretrained visual encoder and finetuning it during language learning both yields better word-learning performance. Similar experiments will be useful to explore how changing the visual encoder can influence the performance of our models.

\section*{Ethics Statement}

We do not anticipate any ethical concerns associated with this work.

\bibliography{ref}

\appendix

\section{Methods}

\subsection{Training and Evaluating Details}\label{ap_train_eval}

\textbf{Network Architecture and Tokenizer.}
We employ a six-layer Transformer network \citep{Transformer} for all our models, featuring per-token hidden representations with a dimension of 768. Each layer is equipped with 12 attention heads, and the feedforward layers post-attention boast an intermediate dimension of 3072. The tokenizer, borrowed from BERT \citep{BERT}, offers a vocabulary size of 30,522. Crucially, we tie the weights of the word-embedding layer with those in the final output layer, a strategy found to be vital for the success of grounded models. For visual encoding, we utilize features extracted via pretrained weights from Huggingface \citep{wolf2020transformers}, specifically using the model ID \verb|facebook/dino-vitb16|.

\begin{figure*}[t]
\centering
\includegraphics[width=0.7\textwidth] {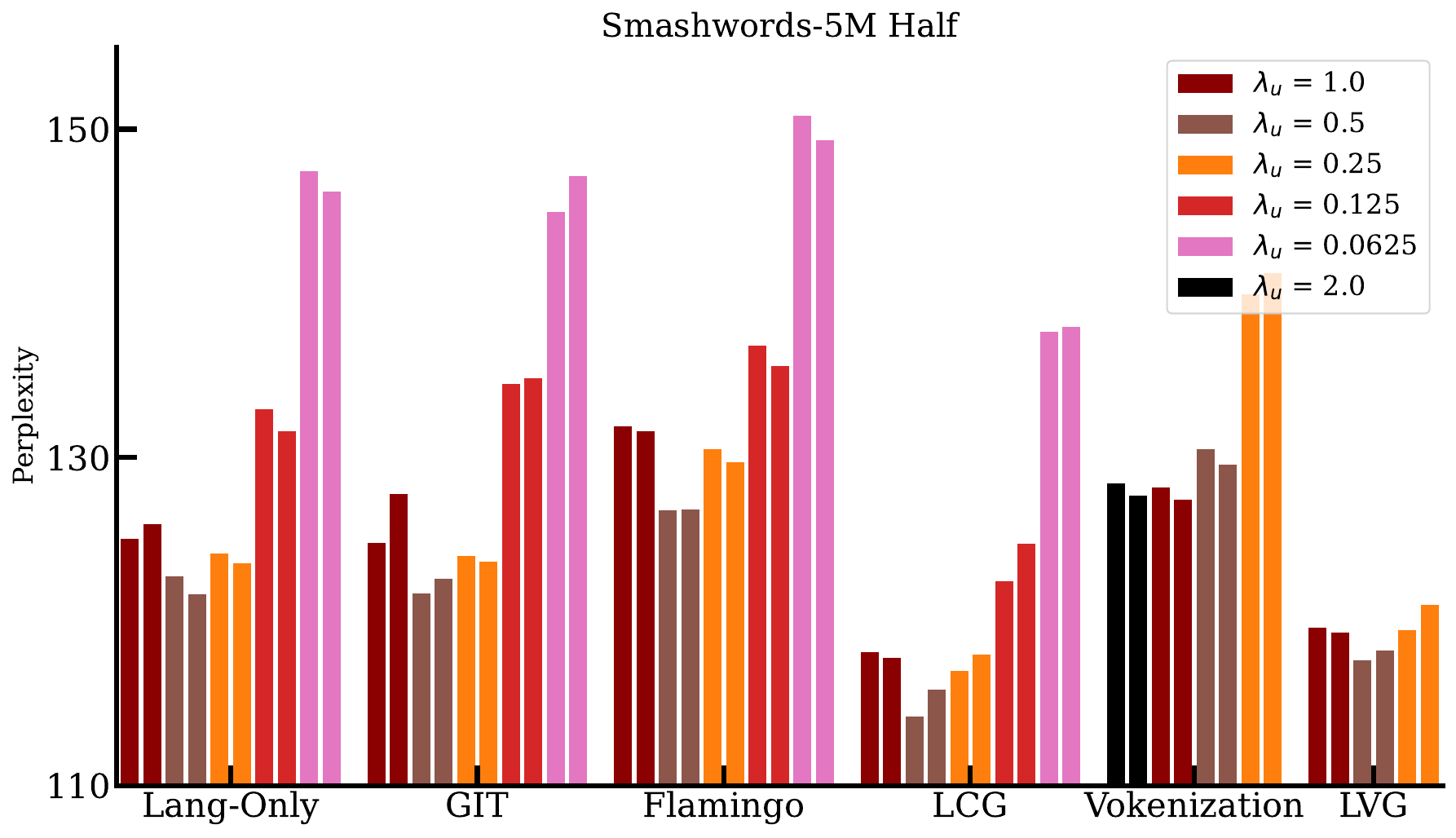}
\captionof{figure}{
\textbf{Perplexity on the Smashwords validation set for models trained with different $\lambda_{u}$ in the training setup with 5M tokens from Smashwords and 2.5M tokens in coupled image-caption pairs.}
For each algorithm and each $\lambda_{u}$, two models are trained from different initialization seeds.
LCG represents the \AlgName, and LVG represents \AlgVokenName.
}
\label{ap_fig_ctr_sw5m_half}
\end{figure*}

\begin{figure*}[t]
\centering
\includegraphics[width=0.7\textwidth] {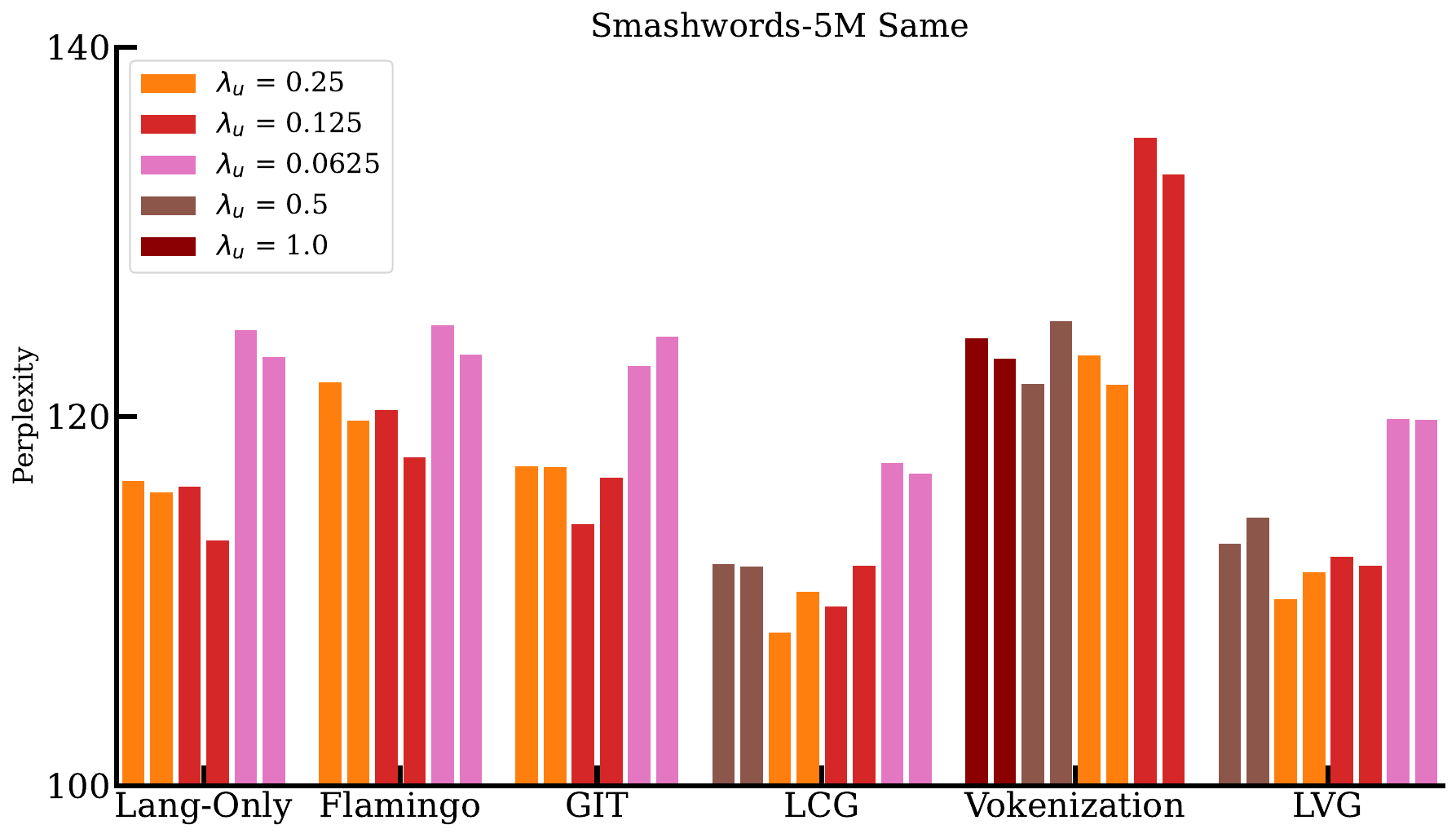}
\captionof{figure}{
\textbf{Perplexity on the Smashwords validation set for models trained with different $\lambda_{u}$ in the training setup with 5M tokens from Smashwords and 5M tokens in coupled image-caption pairs.}
}
\label{ap_fig_ctr_sw5m_same}
\end{figure*}

\begin{figure*}[t]
\centering
\includegraphics[width=0.7\textwidth] {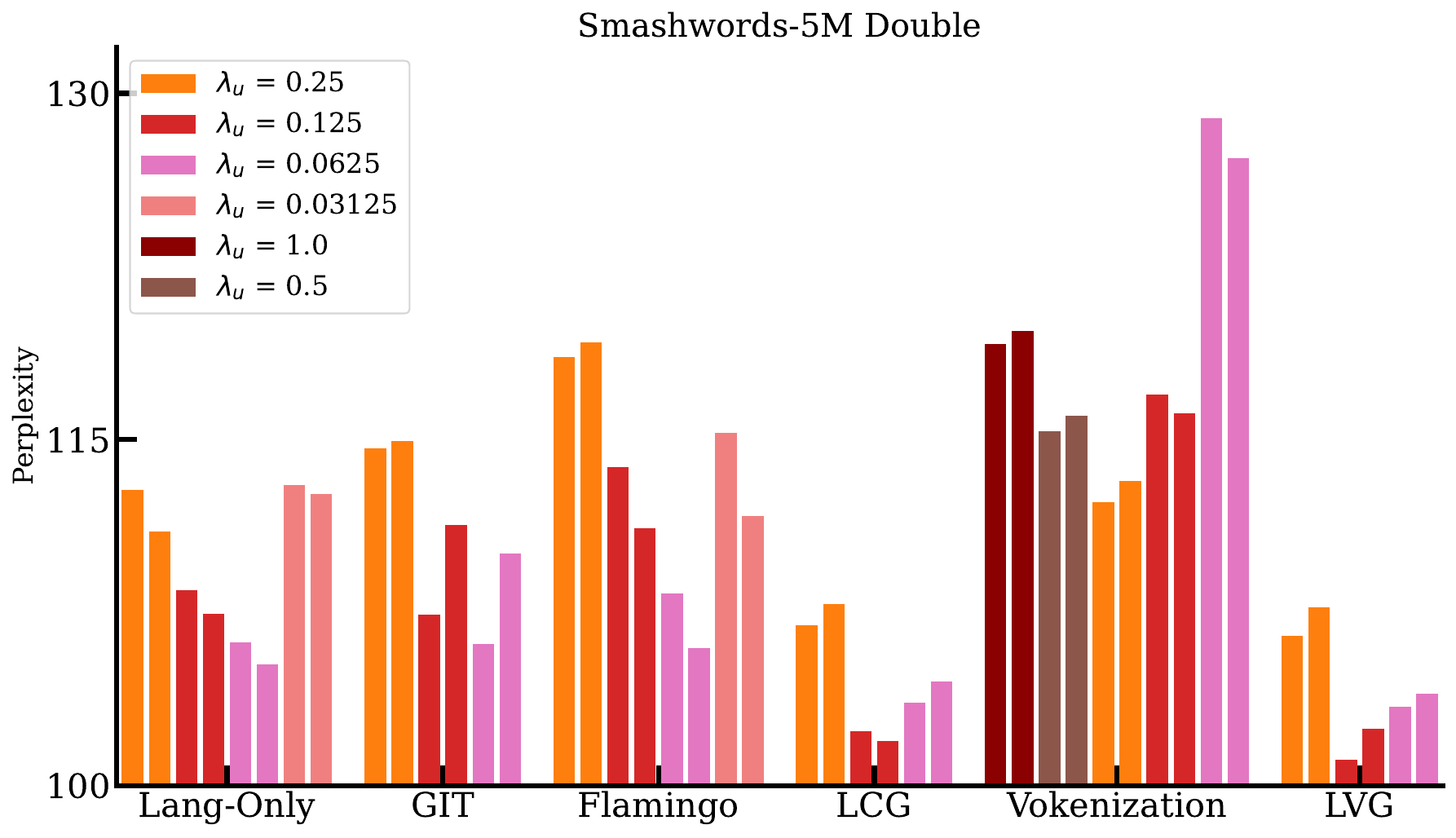}
\captionof{figure}{
\textbf{Perplexity on the Smashwords validation set for models trained with different $\lambda_{u}$ in the training setup with 5M tokens from Smashwords and 10M tokens in coupled image-caption pairs.}
}
\label{ap_fig_ctr_sw5m_double}
\end{figure*}

\begin{figure*}[t]
\centering
\includegraphics[width=0.7\textwidth] {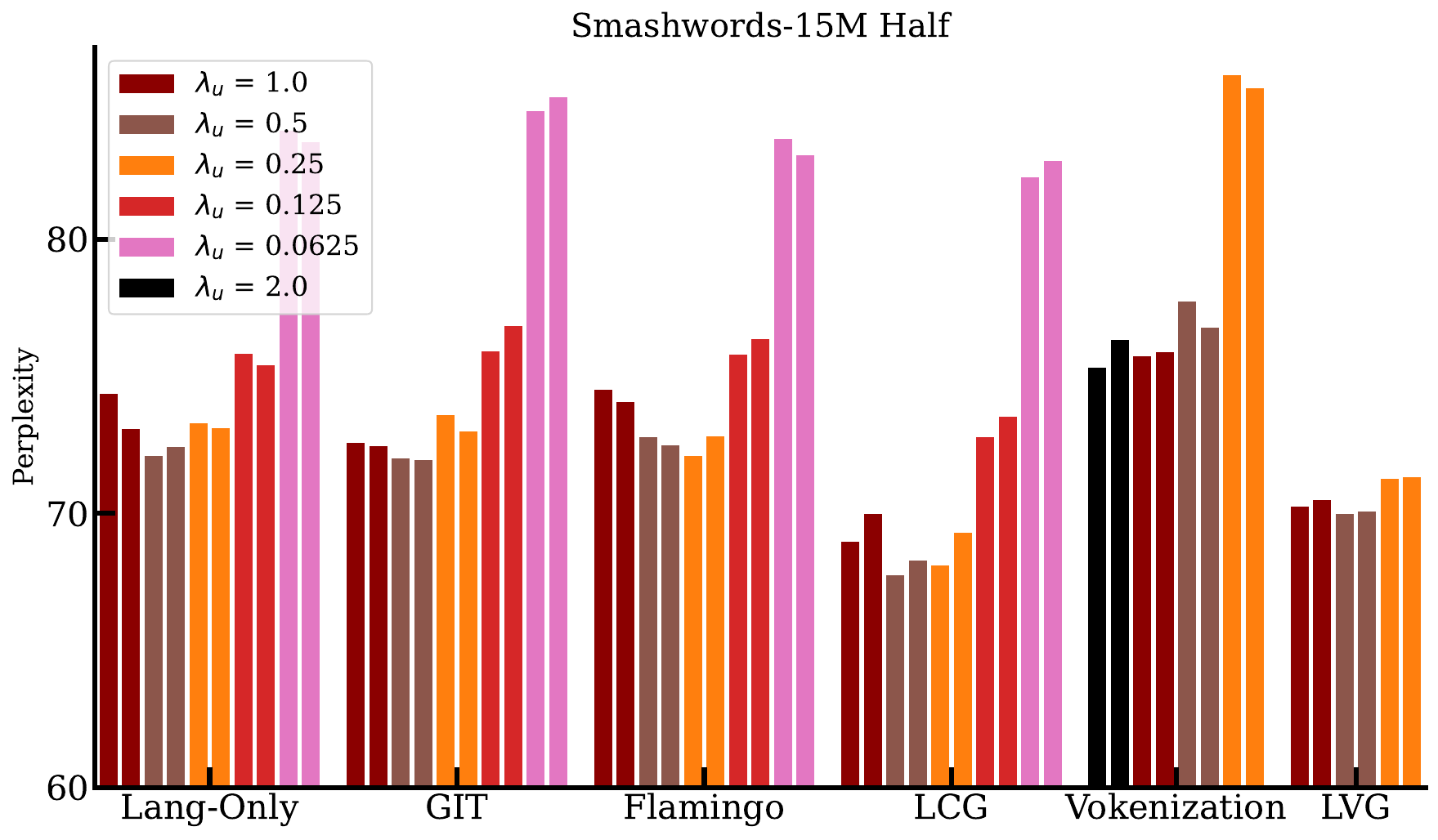}
\captionof{figure}{
\textbf{Perplexity on the Smashwords validation set for models trained with different $\lambda_{u}$ in the training setup with 15M tokens from Smashwords and 7.5M tokens in coupled image-caption pairs.}
}
\label{ap_fig_ctr_sw15m_half}
\end{figure*}

\begin{figure*}[t]
\centering
\includegraphics[width=0.7\textwidth] {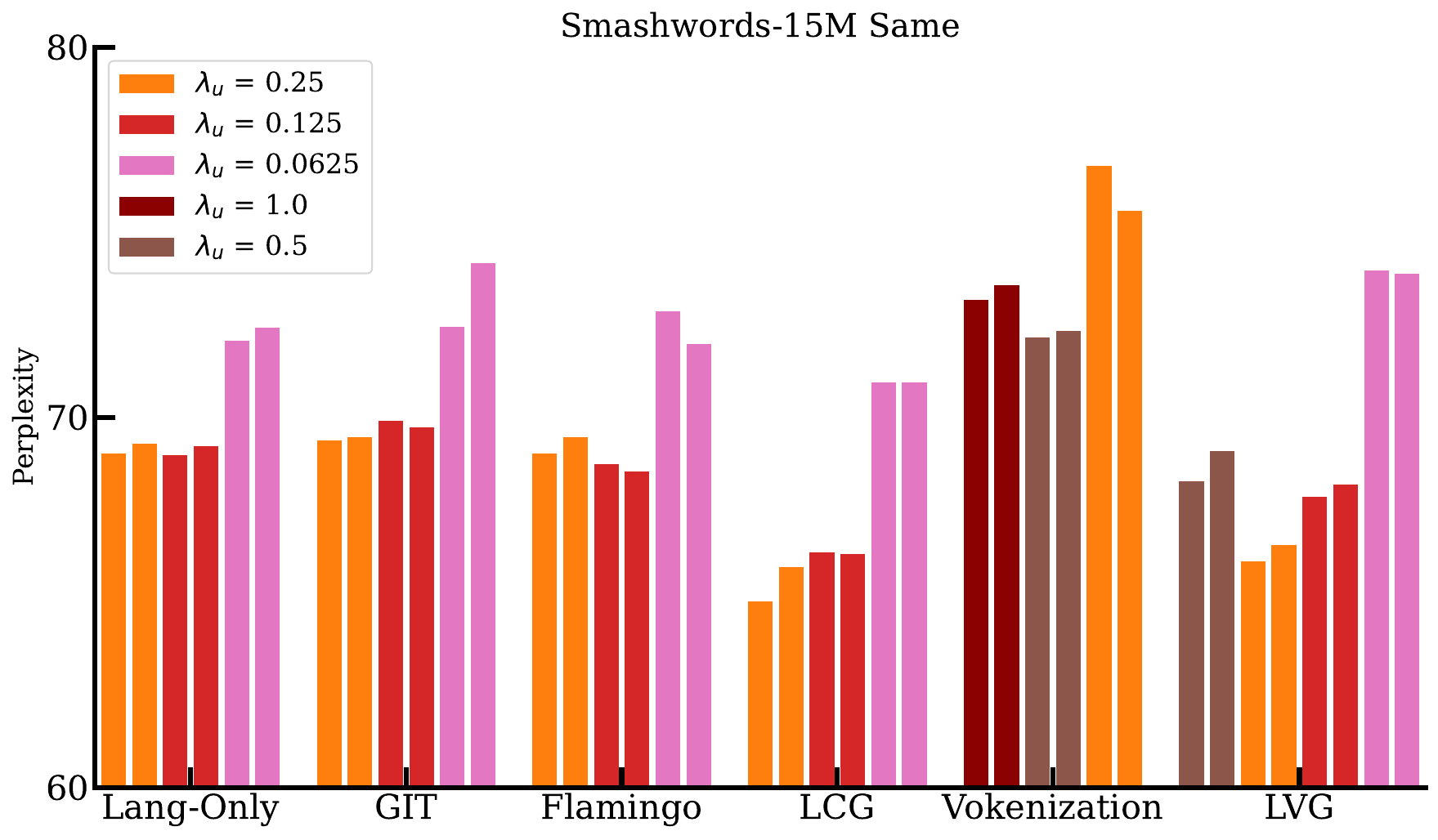}
\captionof{figure}{
\textbf{Perplexity on the Smashwords validation set for models trained with different $\lambda_{u}$ in the training setup with 15M tokens from Smashwords and 15M tokens in coupled image-caption pairs.}
}
\label{ap_fig_ctr_sw15m_same}
\end{figure*}

\begin{figure*}[t]
\centering
\includegraphics[width=0.7\textwidth] {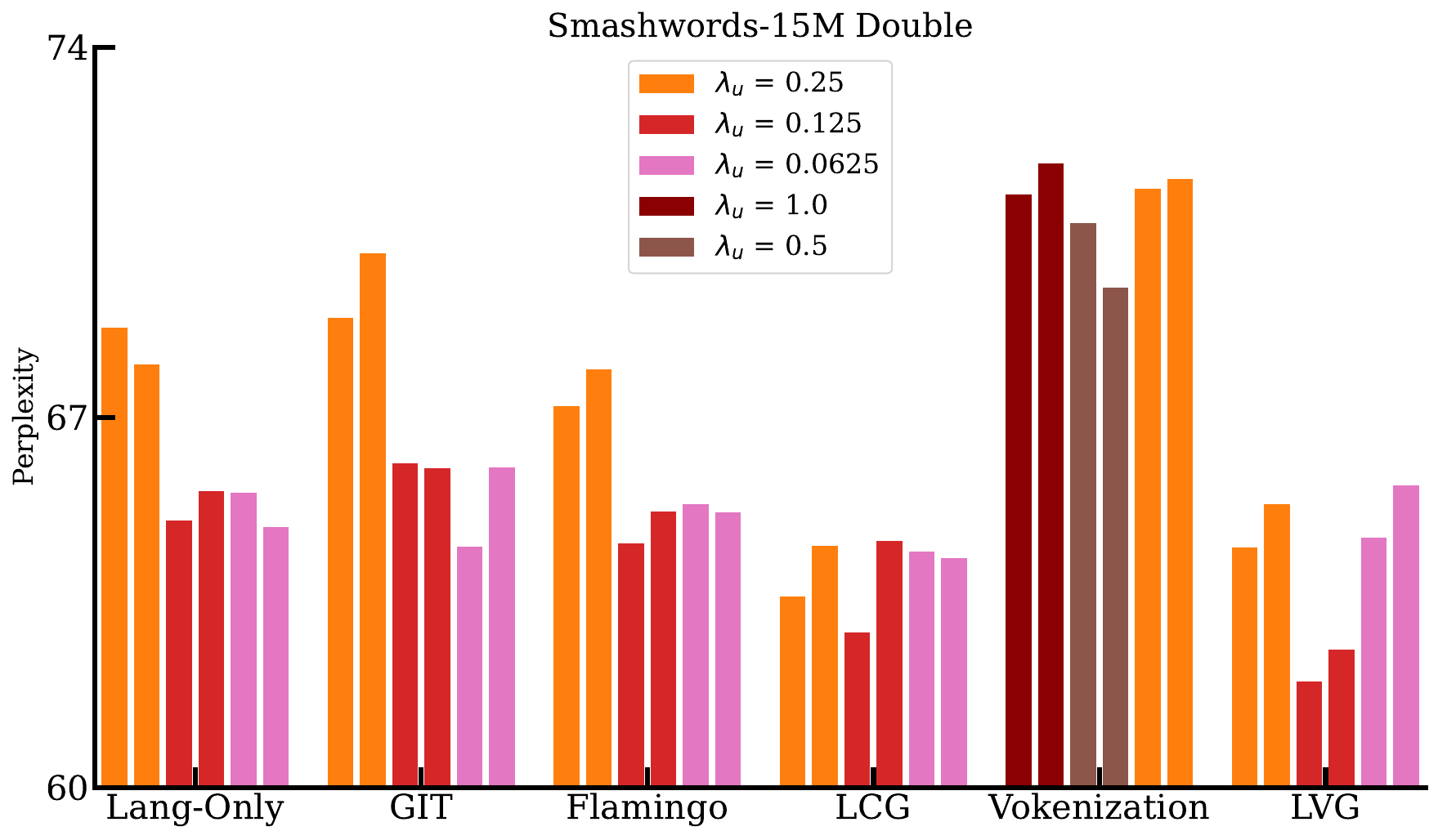}
\captionof{figure}{
\textbf{Perplexity on the Smashwords validation set for models trained with different $\lambda_{u}$ in the training setup with 15M tokens from Smashwords and 30M tokens in coupled image-caption pairs.}
}
\label{ap_fig_ctr_sw15m_double}
\end{figure*}

\begin{figure*}[t]
\centering
\includegraphics[width=0.7\textwidth] {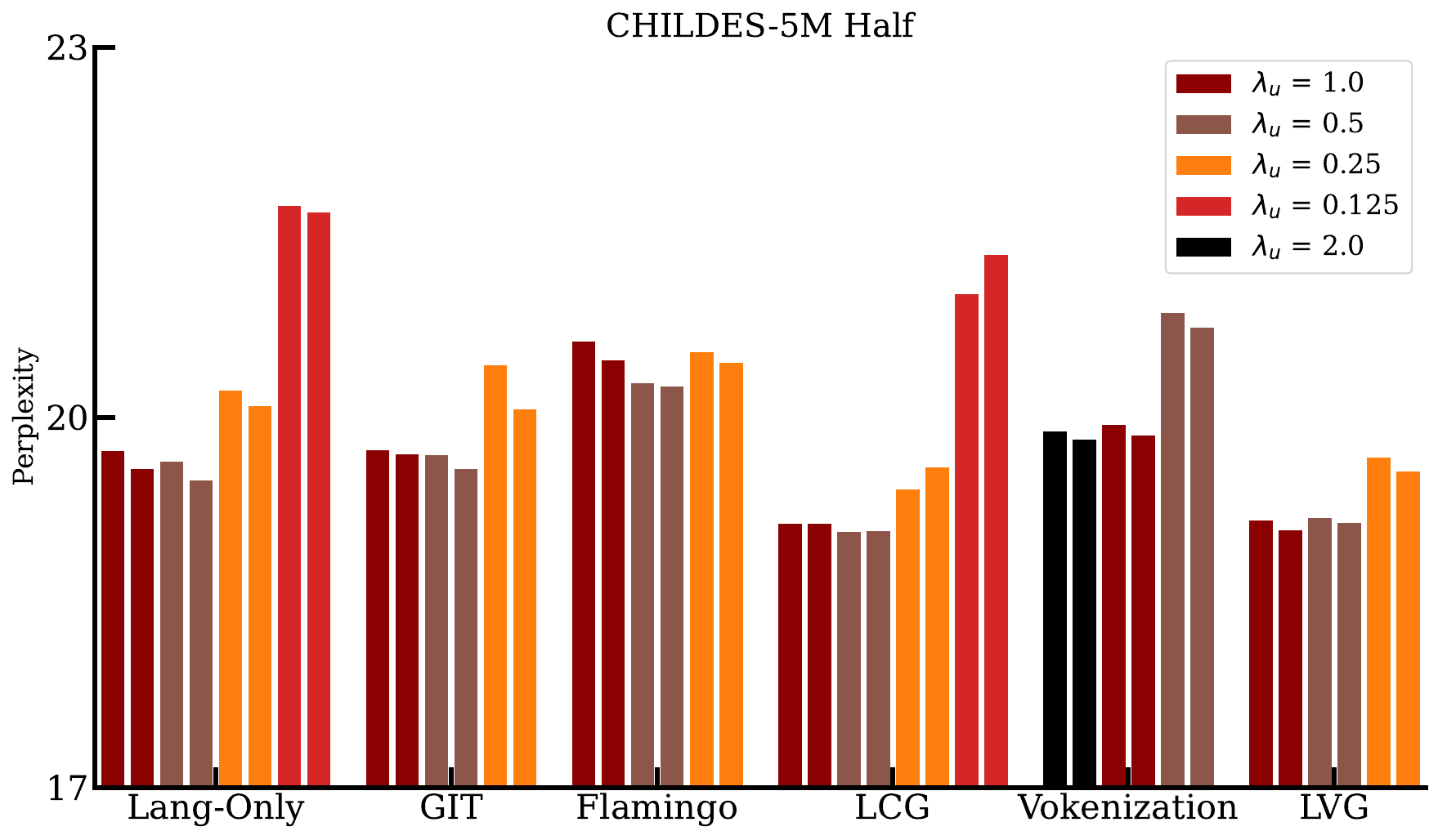}
\captionof{figure}{
\textbf{Perplexity on the CHILDES validation set for models trained with different $\lambda_{u}$ in the training setup with 5M tokens from CHILDES and 2.5M tokens in coupled image-caption pairs.}
}
\label{ap_fig_ctr_chd5m_half}
\end{figure*}

\begin{figure*}[t]
\centering
\includegraphics[width=0.7\textwidth] {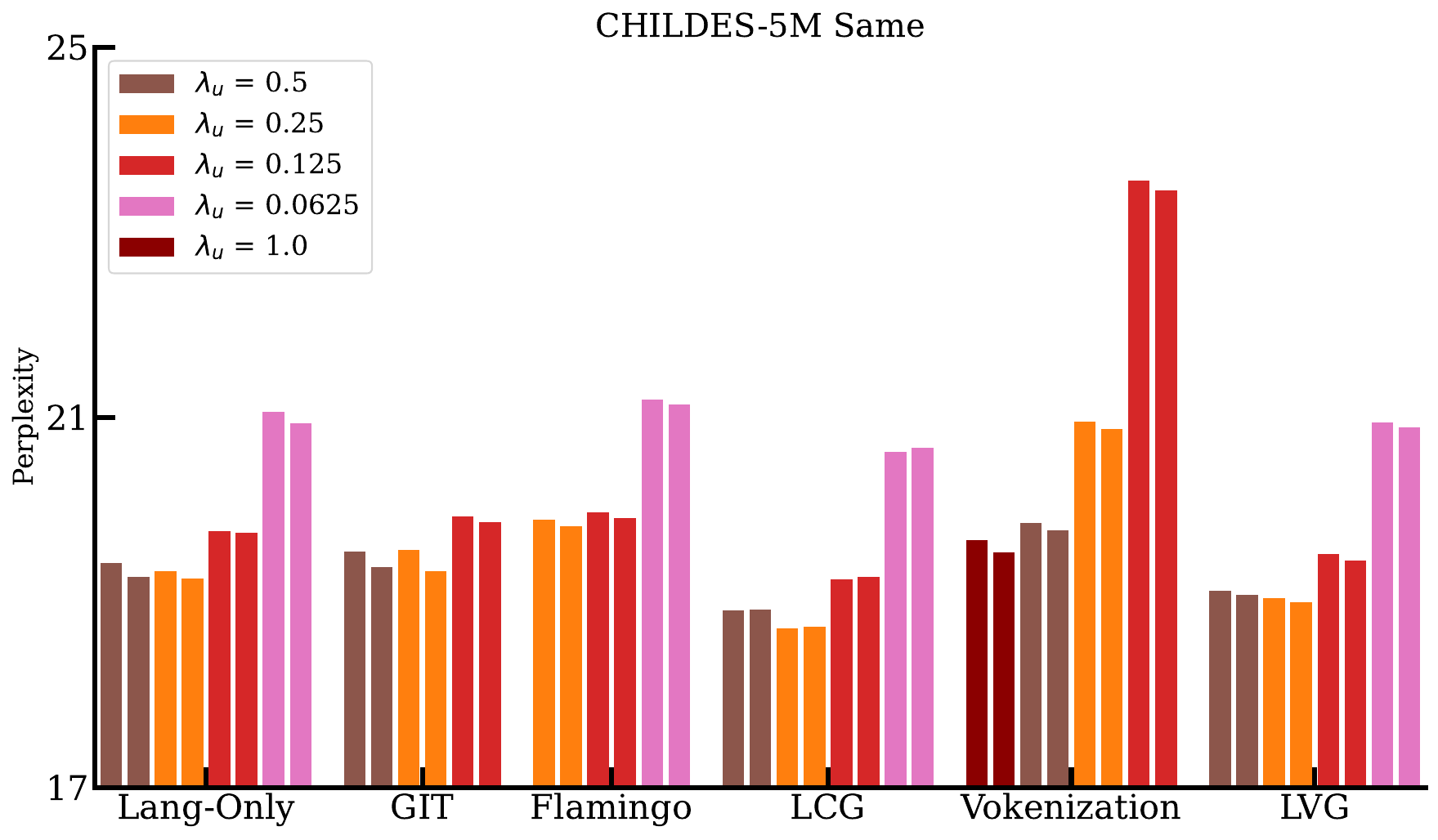}
\captionof{figure}{
\textbf{Perplexity on the CHILDES validation set for models trained with different $\lambda_{u}$ in the training setup with 5M tokens from CHILDES and 5M tokens in coupled image-caption pairs.}
}
\label{ap_fig_ctr_chd5m_same}
\end{figure*}

\begin{figure*}[t]
\centering
\includegraphics[width=0.7\textwidth] {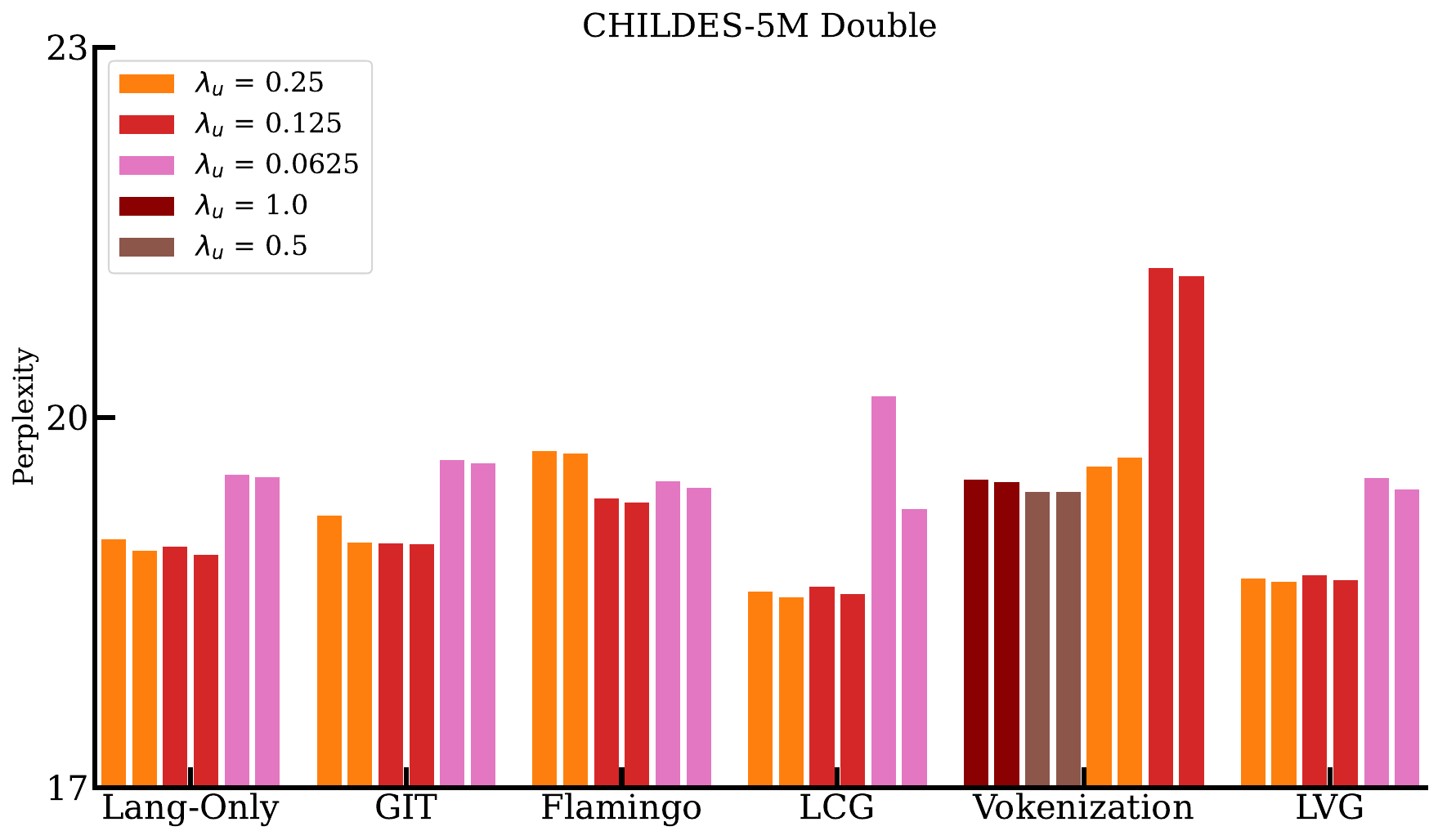}
\captionof{figure}{
\textbf{Perplexity on the CHILDES validation set for models trained with different $\lambda_{u}$ in the training setup with 5M tokens from CHILDES and 10M tokens in coupled image-caption pairs.}
}
\label{ap_fig_ctr_chd5m_double}
\end{figure*}

\textbf{Optimization Details.}
In both grounded-only and mixed-learning scenarios, each model undergoes training across multiple epochs on the datasets.
For grounded-only training, the specific number of epochs is adjusted based on dataset sizes and determined through loss evaluation on the test dataset. 
We use a batch size of 128 for all models except for CLIP, which is trained using a larger batch size of 512.
AdamW \citep{AdamW} is used as the optimizer. 
We initiate the learning rate at zero, incrementing it linearly to 1e-4 over the first 5000 steps, after which it remains constant at 1e-4.
The numbers of epochs for the grounded-only training are as follows:
200 epochs for 100K-token,
40 epochs for 500K-token,
60 epochs for 1M-token,
20 epochs for 5M-token,
and 10 epochs for 15M-token and 50M-token.
These numbers follow the selection by \citet{zhuang2023visual}.
The number of epochs for the mixed training varies independently for different training setups and different algorithms.
These epoch numbers are determined by the perplexity on the validation set of the ungrounded dataset.
In the mixed-learning scenario, the final loss is $\mathcal{L}_{m} = \mathcal{L}_{g} + \lambda_{u} \mathcal{L}_{u}$, where $\mathcal{L}_{u}$ is the loss on the ungrounded batch.
The ungrounded input has a fixed sequence length of 128, a choice inspired by \citet{BabyBERTa}.
For each training setup and each algorithm, we run multiple values for $\lambda_{u}$ and select the best value based on the perplexity on the validation set.
See Fig.~\ref{ap_fig_ctr_sw5m_half} to~\ref{ap_fig_ctr_chd5m_double}.

Our evaluation benchmarks mostly follow the approach of \citet{zhuang2023visual}. Their details are described below.

\textbf{Word Relatedness Benchmark.}
We primarily rely on human assessments of word relatedness gathered by \citet{MTest3000}, where annotators evaluated if one pair of words was more closely related than another. The study focused on words frequently found in both the British Web corpus (ukWaC) and as image tags, resulting in a dataset predominantly consisting of concrete nouns. For the evaluation, each word pair was randomly compared against 50 other pairs, with their relatedness determined by how often they were judged to be more closely related in these comparisons (examples illustrated in Fig.~\ref{fig_main}\textbf{B}). Out of the 3000 word pairs, 2057 pairs were selected for this benchmark to concentrate on words typically learned by children under 10, based on Age of Acquisition metrics from \citet{AoAmetric}. In our models, we measure the relatedness between two words using the cosine similarity of their hidden representations from the same model layer. For words that span multiple tokens, we consider only the representation of the final token. We then calculate Spearman correlations to compare model-derived similarity scores with human relatedness judgments, presenting the highest correlation found across all model layers as the benchmark result for each model.

\textbf{Semantic-Feature Prediction Benchmark.}
We utilize the psycholinguistic feature norms dataset compiled by \citet{Buchanan2019}, where annotators were asked to list any features of a word that came to mind. These responses were processed to isolate single-word features (illustrated in Fig.~\ref{fig_main}\textbf{B}), with the frequency of each feature's occurrence serving as a metric for its significance to the word. The dataset encompasses 3,981 features across 4,436 words. A further selection criterion was applied based on the Age of Acquisition (AoA), restricting the words to those typically learned before the age of 10, which narrowed the list down to 3,554 words. For this benchmark's assessment, we employ a linear regressor trained to estimate a word's feature strength from its hidden model representations. The dataset is divided into training (80\%), validation (10\%), and testing (10\%) segments, with two separate splits created for training and validation to minimize variability. In line with \citet{chronis2023method}, we use a partial least squares (PLS) regressor with 100 components. The evaluation metric is the mean average precision (MAP) across a word's nonzero features, calculated by comparing the top-$k$ predicted features against the actual nonzero features, where $k$ equals the count of ground truth nonzero features. This comparison yields a normalized score based on the overlap. The model layer chosen for hidden representation extraction is determined by its performance on the validation set, with the test set accuracy of this layer then presented as the model's benchmark result.

\textbf{Lexical-Relation Prediction Benchmark.}
The CogALex-V dataset \citep{CogALexV} features 3,054 word pairs in its training division and 4,260 pairs in the testing division. Word pairs with an Age of Acquisition (AoA) exceeding 10 are excluded, resulting in 2,704 pairs for training and 3,900 pairs for testing. A significant portion of these pairs falls under the ``random'' category, accounting for 1,944 of the training and 2,770 of the testing pairs. For model evaluation, we derive the hidden representations of word pairs by calculating the difference between the two words' representations. Consistent with the methodology of \citet{RelBERT}, a Multi-Layer Perceptron (MLP) network is trained to classify lexical relations. We adhere to the standard configurations of the \verb|MLPClassifier| from \verb|sklearn|, observing that adjustments to these settings have a minimal impact on performance. The benchmark's outcome for each model is conveyed through the macro average of F1 scores obtained on the test set by the optimally performing layer.

\textbf{Context Understanding Benchmark.}
This benchmark is constructed by~\citet{zhuang2023visual}.
They selected words known to be learned by young children \citep{WordBank}, covering 140 nouns, 80 verbs, and 60 adjectives.
For each word, they collected example sentences from online websites.
For each example sentence, they then construct minimally different sentences from the example sentence to make the context less appropriate for the original word but more appropriate for distractor words.
This process yields 280K sentence pairs for nouns, 128K for verbs, and 72K for adjectives.

\textbf{Flamingo Training Details.}
The Flamingo model operates with extra cross-attention layers that modulate the outputs from text transformer layers, spaced evenly across the text transformer architecture. 
It processes visual inputs through a Perceiver Resampler equipped with two layers and 64 latents. 
Unlike the visual features in GIT and CLIP models, those in Flamingo include representations from all visual tokens. 
The model, including the Perceiver Resampler, cross-attention, and text transformer layers, was trained from scratch, employing a next-word prediction loss on pairs of images and captions or just sentences.

\textbf{Vokenization Training Details.}
For each training setup in the mixed-learning scenario, we train the contextual token-image matching model on the corresponding image-caption dataset in this setup.
This matching model utilizes the same DINO-pretrained ViT used in other algorithms and the text transformer that is pretrained in this setup, meaning the Language-Only model trained on the corresponding captions and ungrounded dataset.
After training this matching model, we apply it to both the image captions and the ungrounded dataset to map each of the contextual token representations into its most relevant image in a randomly selected subset from VisualGenome, which contains 50K images.
This mapping process yields an index of the chosen image in the range of 0 to 50K, which is the ``voken'' representation corresponding to the token input.
For the Vokenization models, we add an additional readout layer on top of the transformer to predict this voken representation in addition to the standard next-token prediction objective.
For the \AlgVokenName models, the voken prediction is from the output of the first layer in the transformer, which is the same readout location as the \AlgName for visual grounding.

\subsection{Analysis of Learned Representations} \label{ap_analysis}
We follow \citet{zhuang2023visual} to perform the analysis on the results of the word-relatedness benchmark.
In this benchmark, which calculates the Spearman correlation between model outputs and human evaluations, we assign two ranking values to each word pair: one based on the model's assessments and the other on human evaluations.
The absolute discrepancy between these two rankings serves to estimate the model's accuracy in reflecting human-like associations between words. 
To standardize this measure of human likeness, we arrange all word pairs on a scale from least to most human-like—based on the magnitude of their ranking differences. Each pair is then assigned a ``rank in human-likeness,'' where a higher score indicates a closer alignment with human judgment.
The concreteness measure of two words is the average concreteness score of each word~\citep{ConcretenessMetrics}.
A higher concreteness score means that the meaning of the word is more experience-based.

In the analysis for the prediction performance results in Fig.~\ref{fig_analysis}\textbf{C}, we group the words using their concreteness scores into equal-sized five groups.
The negative loglikelihood of a word containing multiple tokens is computed by adding the measure of all tokens together.
We only analyze the words that appear more than five times in the test set.

\subsection{Computational Resources}
We train our models on A100 gpus. Each model has around 70M trainable parameters.
Our implementation majorly uses pytorch and huggingface packages.
The training of all models takes around 2400 GPU hours.

\section{Figures}

\begin{figure*}[t]
\centering
\includegraphics[width=\textwidth] {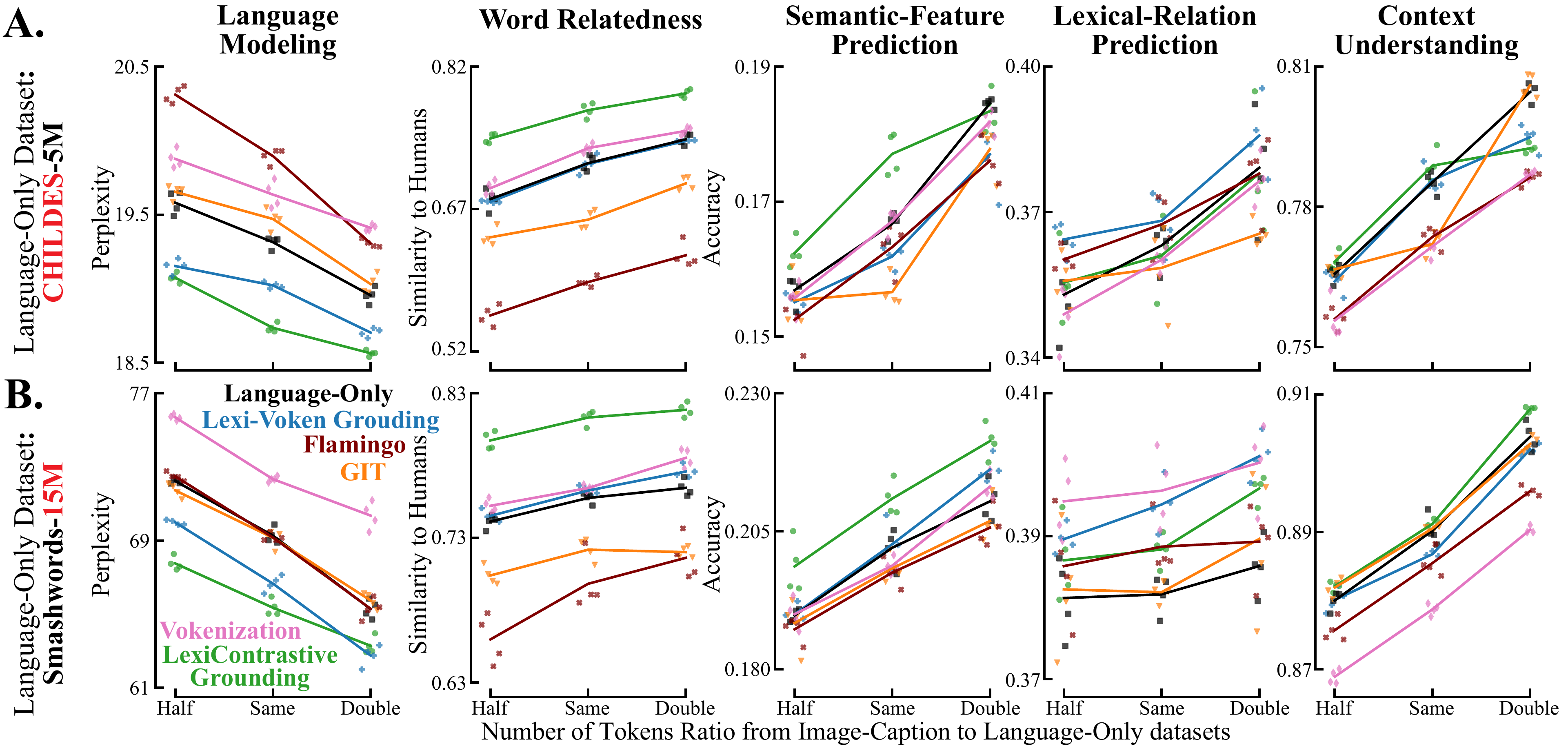}
\captionof{figure}{
\textbf{\AlgName models yield stronger language learning performance than other models when they are co-trained on image-caption and language-only datasets.}
\textbf{A.} Performance on the language modeling and the word learning benchmarks for \AlgName (\ding{108}), \AlgName using Vokens (\ding{58}), Language-Only (\ding{110}), Vokenization (\ding{169}), GIT (\ding{116}), and Flamingo (\ding{54}). The models are trained on a mix of image captions and a language-only dataset containing 5M tokens sampled from CHILDES. The language modeling benchmark evaluates the perplexity of the models on the held-out set of the corresponding language-only datasets. Different dots of the same color represent models with different random initialization seeds.
\textbf{B.} The language-only dataset is a subset of Smashwords containing 15M tokens. 
}
\label{ap_fig_cotrain}
\end{figure*}

\end{document}